# Impacts of Continued Legal Pre-Training and IFT on LLMs' Latent Representations of Human-Defined Legal Concepts

Shaun HO [a,1]

[a] *Carnegie Mellon University*
ORCiD ID: Shaun Ho https://orcid.org/0009-0005-9106-9365

**Abstract.** This paper aims to offer AI & Law researchers and practitioners a more detailed understanding of whether and how continued pre-training and instruction fine-tuning (IFT) of large language models (LLMs) on legal corpora increases their utilization of human-defined legal concepts when developing global contextual representations of input sequences. We compared three models: Mistral 7B, SaulLM-7B-Base (Mistral 7B with continued pre-training on legal corpora), and SaulLM-7B-Instruct (with further IFT). This preliminary assessment examined 7 distinct text sequences from recent AI & Law literature, each containing a human-defined legal concept. We first compared the proportions of total attention the models allocated to subsets of tokens representing the legal concepts. We then visualized patterns of raw attention score alterations, evaluating whether legal training introduced novel attention patterns corresponding to structures of human legal knowledge. This inquiry revealed that (1) the impact of legal training was unevenly distributed across the various human-defined legal concepts, and (2) the contextual representations of legal knowledge learned during legal training did not coincide with structures of human-defined legal concepts. We conclude with suggestions for further investigation into the dynamics of legal LLM training.

## 1. Introduction

The use of LLMs for legal tasks presents challenges because many legal terms bear distinct meanings compared to the same words in general language, and because the statistics of legal corpora differ from those of general corpora [1,2]. While continued pre-training and fine-tuning on legal corpora (collectively, "legal training") has improved performance on legal benchmarks [2,3], whether such models perform better on diverse real-world applications ranging from annotation, rhetorical role prediction, and fact pattern coding [4–7], remains uncertain. This uncertainty is a significant issue for researchers and practitioners since pretraining and fine-tuning data is scarce and costly to obtain, and its quality is highly dependent on annotator expertise and agreement [8–10]. The wider literature has also called into question whether fine-tuning introduces superficial improvements at the cost of model stability [11,12]. A deeper understanding of the underlying dynamics of legal training is required to shed light on the strengths and weaknesses of legal LLMs, so that they may be deployed prudently and appropriately.

---
[1] Corresponding Author: Shaun Ho, jieshauh@andrew.cmu.edu

While attention scores and attention structures (patterns in attention scores along the length of the input sequence) are not completely interpretable [13,14], past examinations provide sufficient evidence that they capture key linguistic features [15–18] and that they contribute meaningfully to output quality [19–23]. Attention scores are also dependent on learned weights which are directly updated during continued pre-training and fine-tuning, thus rendering attention scores a useful indicator of the impact of these processes. Specifically, attention scores are dependent on the degree of overlap between the query vector and the large principal components of the matrix product of the key and query vectors [24], with the projection weights that generate these vectors being altered during continued pre-training and fine-tuning. Finally, analyzing attention scores and structures offers a dataset-independent measure of the extents to which LLMs utilize legal concepts, thus ruling out the influence of dataset quality issues mentioned above.

## 2. Methodology[2]

### 2.1. Models

We compare the general-purpose Mistral 7B[3] [25] with two variants that have undergone continued pretraining and IFT on legal corpora: SaulLM-7B-Base and SaulLM-7B-Instruct [26].[4] The three models utilize identical architectures, thus providing an ideal basis of comparison for isolating the impact of legal training on general-purpose LLMs.

### 2.2. Human-Defined Legal Concepts

Attention heads have been shown to target tokens representing particular parts of speech [15–18]. We assess the extent to which attention heads analogously target tokens in 7 text sequences that represent what will be referred to broadly as "human-defined legal concepts", as shown in Table 1. The legal concepts and text sequences are reproduced directly from recent literature examining LLM abilities to classify each specific text sequence into its corresponding legal concept.

**Table 1.** List of legal concepts, with their corresponding locations within text sequences highlighted in bold.

| Legal Concept | Text Sequence |
|---|---|
| Condition (Pawar et al.) [27] | Section 25F of Industrial Disputes Act: No workman employed in any industry **who has been in continuous service for not less than one year under an employer** shall be retrenched by that employer until... |
| Definiendum (Pawar et al.) [27] | Section 2(s) of Industrial Disputes Act: "**workman**" means any person (including an apprentice) employed in any industry to do any manual, unskilled, skilled, technical, operational, clerical or supervisory work for hire or reward... |
| Evidence Object (Pawar et al.) [27] | Section 25N(7) of IDA: Where no application for permission under sub-section (1) is made, or where the permission for any retrenchment has been refused, such retrenchment shall be deemed to be illegal from the date on which the **notice of retrenchment** was given to the workman … |

---

[2] All code and data will be made available at https://github.com/shaunho7/legal-llm-attention.

[3] Mistral-7B-v0.1.

[4] SaulLM-7B-Base is a variant of Mistral 7B that has undergone continued pretraining on a 30B token dataset comprising primarily of US, EU, and UK legislation and case law. SaulLM-7B-Instruct was obtained by fine-tuning SaulLM-7B-Base on generic instruction datasets as well as a synthetic dataset simulating tasks such as legal classification and summarization.

| | |
|---|---|
| Permissible Action (Pawar et al.) [27] | Section 4(a) of Securitisation Act: ...**take possession of the secured assets of the borrower** including the right to transfer by way of lease, assignment or sale for realising the secured asset |
| Prohibitory Action (Pawar et al.) [27] | Section 3 of Motor Vehicles Act: **No person shall drive a motor vehicle in any public place unless he holds an effective driving licence** issued to him authorising him to drive the vehicle |
| Role (Pawar et al.) [27] | Section 7(1) of Prevention of Money-Laundering Act, 2002: The Central Government shall provide each **Adjudicating Authority** with such **officers** and **employees** as that Government may think fit. |
| Fact Elements (Drápal et al.) [28] | At an undetermined time between 18:00 on May 12, 2017 and 06:00 on May 13, 2017, at the parked delivery vehicle branded Peugeot Boxer, an unknown individual… **entered the vehicle and stole from it** a car radio, a demolition hammer, an electric saw, a drill, and other work tools, all valued at 8,700 CZK … |

*2.3. Computing the Proportion of Attention Directed to Legal Concepts*

To identify parts-of-speech targeted by each attention head, Vig [16] computed the proportion of total attention each head directed toward subsets of tokens representing given parts-of-speech tags. We adapt code used in [15] to compute the same[5] for subsets of tokens representing legal concepts, as demarcated in bold in each item of Table 1.

*2.4. Identifying Legal Features Learned During Legal Training*

Attention patterns in a given attention head have been visualized to determine possible lexical and semantic features captured by that head [16]. To identify legal and semantic features captured during legal training and fine-tuning of Mistral 7B, we visualize patterns of raw attention score differences between Mistral 7B and SaulLM-7B-Instruct.

## 3. Results and Discussion

*3.1. Continued legal pre-training often diminished attention toward legal concepts, while additional legal IFT attenuated these impacts and stabilized the model.*

**Table 2.** Distributions of changes in the proportion of total attention each attention head allocated to legal concepts, grouped by distinct legal concepts and models (SaulLM-7B- "Base" and "IFT"), vs Mistral 7B.

| Legal Concept | Skewness[6] | | Kurtosis[7] | | Entropy[8] | |
|---|---|---|---|---|---|---|
| | Base | IFT | Base | IFT | Base | IFT |
| Condition | –0.33 | **+0.39** | **6.32** | 5.44 | 1.38 | **1.52** |
| Definiendum | –1.69 | **+0.27** | **14.05** | 10.02 | 1.11 | **1.29** |
| Evidence Object | –1.65 | **+0.60** | **9.74** | 7.24 | 1.37 | **1.48** |
| Permissible Action | –0.84 | **–0.16** | **7.37** | 4.08 | 1.38 | **1.71** |
| Prohibitory Action | +1.14 | **+1.33** | **12.35** | 11.43 | 1.27 | **1.29** |
| Role | –0.88 | **–0.08** | **6.85** | 6.80 | **1.40** | 1.37 |
| Fact Elements | –0.77 | **–0.23** | **5.55** | 5.06 | 1.61 | **1.63** |

---

[5] While Vig filters attention to the start-of-sequence token, we additionally filter attention to punctuation.

[6] Skewness measures the asymmetry of a distribution about its mean. Here, negative (positive) values indicate that attention shifts are left (right) skewed toward lower/negative (higher/positive) values.

[7] Kurtosis measures the "tailedness" of a distribution. Higher values indicate "fatter tails" where extreme values are more likely to be observed. Here, higher values imply more attention shifts in the extremes.

[8] Entropy measures the "disorder" of a distribution. Higher values indicate a more even spread of values. Entropy was computed using 11 bins in accordance with Sturges' formula ($n$ = 32 layers × 32 heads = 1024).

*Skewness.* Attention shifts brought about by continued pre-training of Mistral 7B demonstrated moderate to strong negative skewness in most instances, suggesting that continued pre-training on legal corpora diverted attention *away* from most human-defined legal concepts, sometimes to an extreme extent. Additional IFT modulated this behavior, resulting in skewness values that were consistently closer to zero or even positive (i.e. directing additional attention *toward* human-defined legal concepts).

*Kurtosis and Entropy.* The distribution of attention shifts toward legal concepts exhibited lower kurtosis after IFT, further suggesting that IFT modulated instabilities introduced during continued pre-training. Entropy was slightly higher following IFT. When taken together with reduced kurtosis, this indicates that IFT distributed the effects of legal training more uniformly (across a slightly broader range of attention heads).

*3.2. Legal training unevenly impacted the model's utilization of distinct legal concepts when developing global contextual representations of input sequences.*

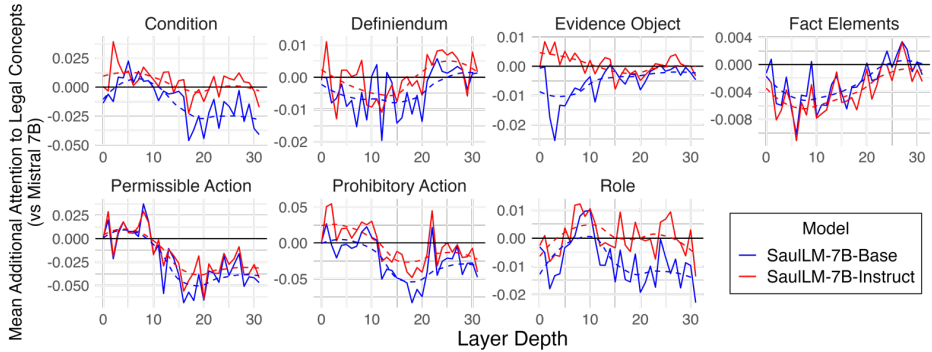

**Figure 1.** Additional proportion of total attention SaulLM-7B (-Base and -Instruct) allocated to tokens representing human-defined legal concepts (vs Mistral 7B); mean increase across all heads in each layer.

Notably, neither continued pretraining nor fine-tuning on legal data resulted in a discernible difference in model attention to many human-defined legal concepts (*Fact Elements, Evidence Object, Definiendum, Role*). The mean values of attention shifts toward these legal concepts rarely fell outside the range of [–0.01, 0.01].

In some cases (*Condition, Evidence Object, Permissible Action, Prohibitory Action*), the proportion of attention directed toward legal concepts was increased slightly in earlier layers, especially after IFT. This was often followed by a diversion of attention away from legal concepts in the later layers (see also *Role*). Previous work examining attention structures by layer depth provides insight into this observation. It has been demonstrated that initial attention layers attend more broadly and capture higher-level concepts, while subsequent layers attend more narrowly to lower-level concepts [16,21,29,30]. Legal training therefore (1) improved Mistral 7B's utilization of human-defined legal concepts when developing broader, longer-ranged, and higher-level contextual representations of input sequences, but (2) often diminished its reliance on legal concepts when developing narrower, recent-ranged, and lower-level representations of those input sequences.

In most cases the mean proportion of total attention directed toward legal concepts was almost always more positive following additional IFT, regardless of layer depth. This suggests that the improvements brought about by additional IFT were evenly distributed and reflected across all levels of the feature hierarchy.

### 3.3. Legal training did not introduce novel attention structures of legal knowledge.

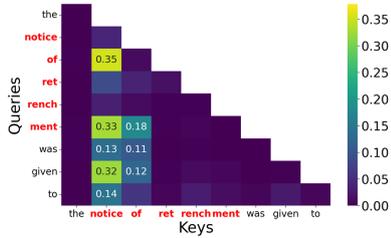

**Figure 2.** Mistral 7B attention matrix in Layer 3, Head 31 on the text sequence "Evidence Object".

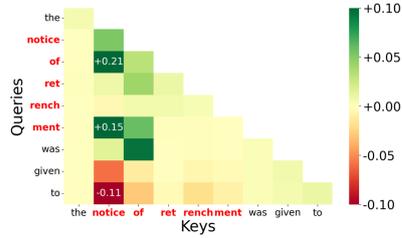

**Figure 3.** Differences in raw attention scores between SaulLM-7B-Instruct and Mistral 7B.

Visualizations revealed that the patterns of alterations learned during legal training (Figure 3) targeted pre-existing attention structures in Mistral 7B (Figure 2), rather than introducing new structures corresponding to human-defined legal concepts.

Legal feature representation was also degraded after legal training and IFT. Mistral 7B directed significant attention to `notice` from the queries `was`, `given`, and `to` (Figure 2), possibly capturing an essential aspect of effective notice pertaining to its service on a person (*Mullane v. Central Hanover Bank & Trust Co.*, 339 U.S. 306 (1950)). In contrast, Saul-LM-Instruct counter-productively suppressed these relations (Figure 3).

Furthermore, legal training did not introduce patterns corresponding to human-defined legal concepts. Figure 3 indicates that legal training altered attention to `notice` and `of`, but not `ret`, `rench`, or `ment`. This does not satisfy the intention of Pawar et al. [27] to treat the entire *contiguous sequence* of tokens `notice of retrenchment` as a single class object. Likewise, attention toward `jud` was increased in Layer 28, Head 16 but not toward the other tokens in the set of `Adjudicating Authority`.

### 4. Conclusion and Future Work

These preliminary results identify the need for further investigation (across a broader range of models, input sequences, legal concepts, and jurisdictions) into the following:

*Uneven effects of legal training.* Continued legal pretraining often suppressed attention to human-defined legal concepts, with IFT attenuating these effects. There were also inconsistencies in how far legal LLMs utilized the different legal concepts when developing global contextual representations of input sequences. This demonstrates that the effectiveness of legal LLMs varies with the specific legal concepts used in each task.

*LLMs' contextual representations of legal knowledge do not align with human representations.* A key insight is that while human-defined legal concepts are often represented by contiguous sequences of tokens, attention patterns learned during legal training often only target a small subset of tokens within those sequences. This may call for new tokenization strategies, such as that of Nguyen et al. [31] which improved legal task performance by using custom tokens to highlight legal keywords in input sequences.

*The influence of base models and architectures on legal training.* The results in this paper support the argument that instead of developing novel capabilities, "fine-tuning protocols merely identify the most relevant capabilities and amplify their use for a given set of inputs" [12]. Further work should explore the extent to which the choice of base model (e.g. Mistral 7B) influences the outcomes of continued legal pre-training and IFT.